\pdfoutput=1

\documentclass[11pt]{article}

\usepackage[]{acl}

\usepackage{times}
\usepackage{latexsym}
\usepackage{CJKutf8}
\usepackage[T1]{fontenc}

\usepackage[utf8]{inputenc}

\usepackage{microtype}

\usepackage{graphicx}
\usepackage{multirow}
\usepackage{amsmath,amsfonts,amssymb}
\usepackage{algorithmic}
\usepackage[ruled,vlined]{algorithm2e}

\DeclareMathOperator{\E}{\mathrm{E}}
\DeclareMathOperator{\Var}{\mathrm{Var}}

\usepackage{bm}
\usepackage{array}

\newcommand{\cmtt}[1]{{\fontfamily{cmtt}\selectfont {#1}}}
\usepackage[colorinlistoftodos,prependcaption,textsize=scriptsize]{todonotes}

\DeclareFontShape{OT1}{cmtt}{bx}{n}
{
<5> <6> <7> <8> <9>
<10> <10.95> <12> <14.4> <17.28> <20.74> <24.88> cmbtt10
}{}
\DeclareFontShape{OT1}{cmtt}{b}{n}
{<->sub * cmtt/bx/n}{}

\title{Disentangled Sequence to Sequence Learning for \\ Compositional Generalization }

\author{Hao Zheng \textnormal{and} Mirella Lapata\\
Institute for Language, Cognition and Computation\\
School of Informatics, University of Edinburgh\\
 10 Crichton Street, Edinburgh EH8 9AB\\
\texttt{Hao.Zheng@ed.ac.uk}~~~~\texttt{mlap@inf.ed.ac.uk}\\
}

\date{}

\makeatletter
\newcommand{\thickhline}{%
    \noalign {\ifnum 0=`}\fi \hrule height 1pt
    \futurelet \reserved@a \@xhline
}
\makeatother

\begin{document}
\maketitle
\begin{abstract}
  There is mounting evidence that existing neural network models, in
  particular the very popular sequence-to-sequence architecture,
  struggle to systematically generalize to unseen compositions of seen
  components.  We demonstrate that one of the reasons hindering
  compositional generalization relates to representations
  being \emph{entangled}. We propose an extension to
  sequence-to-sequence models which encourages disentanglement by
  adaptively re-encoding (at each time step) the source
  input. Specifically, we condition the source representations on the
  newly decoded target context which makes it easier for the encoder
  to exploit specialized information for each prediction rather than
  capturing it all in a single forward pass. Experimental results on
  semantic parsing and machine translation empirically show that our
  proposal delivers more disentangled representations and better
  generalization. \footnote{Our code is available at \url{https://github.com/mswellhao/Dangle}.}

\end{abstract}

\section{Introduction}
\label{sec:introduction}

When humans use language, they exhibit compositional generalization;
they are able to produce and understand a potentially infinite number
of novel linguistic expressions by systematically combining known
atomic components \cite{Chomsky, montague1970universal}. For example,
if a person knows the meaning of the utterance ``A boy ate the cake on
the table in a house'' and the verb ``like'', it is natural for them
to understand the utterance ``A boy likes the cake on the table in a
house'' when they encounter it for the first time (see
Table~\ref{fig:split}).  Humans are also adept at recognizing novel
combinations of familiar syntactic structure, e.g., they would have no
trouble processing the above sentence if the preposition ``beside the
tree'' were added to it, despite not having previously seen the phrase
``in a house beside the tree'' (see Table~\ref{fig:split}).

\begin{table}[t]
\centering
\small
\begin{tabular}{@{}p{7.8cm}@{}} \hline
  \multicolumn{1}{c}{\em Training Set} \\ \hline
  A boy ate the cake on the table in a house. \\
 *cake($x_4$); *table($x_7$); boy(x$_1$) AND eat.agent(x$_2$, x$_1$) AND
  eat.theme(x$_2$, x$_4$) AND cake.nmod.on(x$_4$, x$_7$) AND
  table.nmod.in(x$_7$, x$_{10}$) AND house(x$_{10}$)  \\ \hline  
  \multicolumn{1}{c}{\em Test Set (Lexical Generalization)} \\ \hline
  A boy likes the cake on the table in a house. \\
  *cake(x$_4$); *table($x_7$); boy(x$_1$) AND like.agent(x$_2$, x$_1$) AND
  like.theme(x$_2$, x$_4$) AND cake.nmod.on(x$_4$, x$_7$) AND
  table.nmod.in(x$_7$, x$_{10}$) AND house(x$_{10}$) \\ 
  \hline
  \multicolumn{1}{c}{\em Test Set (Structural Generalization)} \\ \hline
  A boy ate the cake on the table in a house beside the tree. \\ 
  *cake(x$_4$); *table(x$_7$); *tree(x$_{13}$); boy(x$_1$) AND eat.agent(x$_2$,
  x$_1$) AND eat.theme(x$_2$, x$_4$) AND cake.nmod.on(x$_4$, x$_7$) AND
  table.nmod.in(x$_7$, x$_{10}$) AND house(x$_{10}$) 
  AND house.nmod.beside(x$_{10}$, x$_{13}$) \\ \hline 
\end{tabular}
\caption{Examples  from COGS  \cite{kim-linzen-2020-cogs}
  showcasing lexical and structural  generalization. 
  In lexical generalization, a familiar word (e.g.,~\textit{like}) is
  attested in a familiar syntactic structure  but the resulting
  combination has not been seen before. In structural generalization,
  familiar syntactic components give rise to novel combinations
  (e.g.,~only prepositional phrases with nesting depth~2 have been
  previously seen whereas new combinations show nestings of depth~3
  or~4). All PP   modifiers are assumed to  have an NP-attachment reading and all modifications are nested 
  rather than sequential. Definite descriptions  are marked with~* and
  appear to the leftmost of the logical form.}

\label{fig:split}
\end{table}

There has been a long standing debate whether this systematicity can
be captured by connectionist architectures
\cite{fodor1988connectionism, Marcus2003,lake2018generalization} and
recent years have witnessed a resurgence of interest thanks to the
tremendous success of neural networks at various natural language
understanding and generation tasks \cite{NIPS2014_a14ac55a, NIPS2017_7181,dong-lapata-2016-language,jia-liang-2016-data}.  Mounting evidence, however, suggests that existing models, in particular the very popular sequence-to-sequence architecture, struggle with compositional generalization
\cite{finegan-dollak-etal-2018-improving,lake2018generalization,keysers2020measuring,herzig2020spanbased}.  This failure may be due to spurious
correlations which hinder out-of-distribution generalization
\cite{gururangan-etal-2018-annotation,arjovsky2019invariant,Sagawa*2020Distributionally}
or limited robustness to perturbations in the input
\cite{cheng-etal-2018-towards}.


In this paper, we identify an
\emph{entanglement problem} with how different semantic factors (e.g.,~lexical meaning and semantic
relations) are represented in neural sequence models that hurts
generalization. In theory, neural networks should represent semantic
factors in a disentangled way by virtue of the principle of
compositionality \cite{frege:1884,Partee:1995} which implies 
that semantic properties of syntactic constituents are to a certain extent
context invariant and the semantic primitives they express are
conditionally independent.

Disentangled meaning representations ought to preserve
this conditional independence, and neural units modeling a particular
semantic factor should be relatively invariant to changes in other
factors \cite{bengio2013representation}. For example, the relation
between ``table'' and ``house'' in Table~\ref{fig:split} and its
representation should not be affected by whether there is a PP
modifying ``house''. However, in a standard neural encoder
(e.g.,~transformer-based) semantic factors tend to be entangled so
that changes in one factor affect the representation of others. We
further illustrate this problem in an artificial setting and find that
a simple marking strategy enhances the learning of disentangled
representations. 

Motivated by this finding, we propose an extension to
sequence-to-sequence (seq2seq) models which allows us to learn
disentangled representations for compositional
generalization. Specifically, at each time step of the decoding, we
adaptively re-encode the source input by conditioning the source
representations on the newly decoded target context. We therefore
build specialized representations which make it easier for the encoder
to exploit relevant-only information for each prediction.  Experiments
on three benchmarks, namely COGS
\cite{kim-linzen-2020-cogs}, CFQ
\cite{keysers2020measuring}, and CoGnition
\cite{li-etal-2021-compositional}, empirically verify that our
proposal leads to better generalization, outperforming competitive
baselines and more specialized techniques.


\section{Disentanglement in a Toy Experiment}
\label{toy_section}

We first shed light on the problem of entangled representations with a
toy experiment and then move on to describe our modeling
solution. For simplicity, we only focus on relations as the kind of semantic factors a
model aims to represent, but the entanglement issue could also exist
in representations of other factors, such as lexical meaning.

\paragraph{Data Creation} Let $x = [e_1, r_1, e_c, r_2, e_2]$ denote a
sequence of symbols. We want to predict the relation between $e_1$ and
$e_c$, and $e_c$ and $e_2$, which we denote by $y=(y_1, y_2)$, with
$y_1 \in L_1$ and $y_2 \in L_2$ where $L_1$ are a set of relation
labels for $y_1$ and $L_2$ are a set of relation labels for $y_2$.
For simplicity, we set $e_1$, $e_c$, and $e_2$ to the same symbol~$e$
(i.e.,~$e_1, e_c, e_2 \in \{e\}$) whereas $r_1 \in R_1$ and $r_2 \in
R_2$ denote different relation symbols, and $R_1$ and $R_2$ are the
corresponding sets of relation candidates. In this toy setting, we
will further assume that different relation symbols determine
different relation labels (e.g., for the phrases ``cat in house'' and
``cat with house'', ``in'' and ``with'' represent two distinct
relations between ``cat'' and ``house'').  In reality, relations
between words could be dependent on broader context or not verbalized
at all. We also assume that there is a one-to-one mapping between
relation symbols and relation labels (i.e.,~between $L_1$ and $R_1$
and $L_2$ and~$R_2$).

We construct a training set by including examples $[e_1, r_1, e_c,
r_2, e_2]$ where $r_1$ is the same relation symbol throughout
while~$r_2$ can be any relation symbol in $R_2$ ($r_1 \in
\{r_{train}\}$, $r_2 \in R_2$). We also include examples $[e_1, r_1,
e_c]$ with all relation symbols from $R_1$ occurring in isolation
$(r_1 \in R_1)$. This way, the training set covers all primitive
relations, but contains only a particular type of relation composition
(i.e., $\{r_{train}\} \times R_2$). In contrast, the test set contains
all unseen compositions $[e_1, r_1, e_c, r_2, e_2]$ (i.e.,~$r_1 \in
R_1 \backslash \{r_{train}\}, r_2 \in R_2$) which will allow us to
evaluate a model's ability to generalize. We set each relation set to
include 10~relation symbols ($|R_1|=|R_2|=$10).

Finally, we simplistically only consider the relations of target
word~$e_c$ with its left and right words~$e_1$ and~$e_2$. In reality,
a model would be expected to capture sentence-level semantics, i.e.,~a
word's relation to \emph{all} context words in a sentence (including
no relation).



\paragraph{Modeling} For each input symbol, we sample a vector from a
Gaussian distribution $\mathcal{N}(\bf{0},0.2^{2}\bf{I})$ and freeze
it during training.  We then embed each example $x$~into a sequence of
vectors $[w_1,w_2,...,w_n]$ (where $n=3$ or $n=5$) and  transform
them into contextualized representations $[h_1, h_2,..,h_n]$ using a
Transformer encoder \cite{NIPS2017_7181}. To predict the relation
between two symbols, we concatenate their corresponding
representations and feed the resulting vector to an MLP for
classification.

To study how changes in relation~$y_1$ affect the prediction of $y_2$
at test time, we explore two training methods. One is joint training
where a model learns to predict both~$y_1$ and~$y_2$ (i.e.,~$h_1$
and~$h_3$ are concatenated to predict $y_1$ or $h_3$~and~$h_5$ are
concatenated to predict~$y_2$). The other method is separate training
where a model is trained to only predict~$y_2$ (i.e., only~$h_3$ and
$h_5$~are concatenated to predict~$y_2)$. For separate training, we
basically ignore examples $[e_1, r_1, e_c]$ which only include~$r_1$,
as they have no bearing on the prediction of~$y_2$.

\paragraph{Observation} With separate training, the model learns to
ignore~$r_1$, the accuracy of predicting~$y_2$ on the test set is
100\%, regardless of which value $r_1$~takes. This indicates that
random perturbation of~$r_1$ alone does not lead to generalization
failure. It also follows that there is no spurious correlation
between~$r_1$ and~$y_2$. However, when the model is trained to predict
both relations (which is what happens in realistic settings since we
need to capture all possible relations) $r_1$~has a huge impact on the
prediction of~$y_2$ whose accuracy drops to approximately~55\%. Taken
together, these results suggest that the model fails to generalize to
new relation compositions due to its internal representations being
entangled and as a result changes in one relation affect the
representation of others.


Why is there a wide performance gap between joint and separate
training? At test time the model processes the same utterance (no
matter whether it is trained jointly or separately), and could in
theory be susceptible to both $r_1$ and $r_2$. However, the induced
representations show fundamentally different behaviors, and remain
invariant to~$r_1$ with separate training. A possible explanation is
that modern neural networks trained with SGD have a learning bias
towards \emph{simple} functions
\cite{NEURIPS2020_6cfe0e61}. When~$r_1$ is not predictive of~$y_2$,
relying only on~$r_2$ whilst remaining invariant to~$r_1$ constitutes
a simpler function than making use of both~$r_1$ and~$r_2$. As a
result, in separate training the model learns to ignore extraneous
information, focusing exclusively on~$r_2$. On the contrary, in joint
training the target of predicting both $y_1$ and $y_2$ forces the
hidden states (e.g., $h_3$) to capture information about both
relations, leading to the entanglement problem discussed above.

\paragraph{A Simple Solution} Although separate training presents a
solution to entanglement, it is unrealistic for real-wold data as it
would be extremely inefficient to train separate models for each
relation (the number of relations is quadratic with respect to
sentence length). Instead, we explore a simple but effective approach
where a single model takes as input an utterance enriched with
different indicator features for different targets.  Specifically,
given utterance $[e_1, r_1, e_c, r_2, e_2]$, and assuming we wish to
predict relation~$y_1$, we add indicator feature~1 for symbols~$e_1$,
$r_1$, and $e_c$ (marking the relation and its immediate context), and
0~for all other symbols. The model then takes as input the utterance
\emph{and} relation indicators, i.e., $[1,1,1,0,0]$ for $y_1$ and
$[0,0,1,1,1]$ for $y_2$, and learns embeddings for indicators during training.
It thus learns specialized representations
for \emph{each} prediction rather than shared representations for
\emph{all} predictions. Based on the simplicity bias, the
two representations will guide the model towards exclusively relying
on~$r_1$ and~$r_2$, naturally disentangling different relations by
encoding them separately. Such a model predicts $y_{1}$ with~100\%
test accuracy and $y_2$ with~97\%.

\paragraph{Discussion}

\citet{fodor1988connectionism} have argued that failure to capture
systematicity is a major deficiency of neural architectures,
contrasting human learners who can readily apply known grammatical
rules to arbitrary novel word combinations to individually memorizing
an exponential number of sentences.  However, our toy experiment shows
that neural networks are not just memorizing sentences but implicitly
capturing structure. With separate training or joint training enhanced
with the marking strategy, the neural model manages to remain robust
to interference from~$r_1$ and properly represent~$r_2$ even for
unseen examples, i.e.,~new compositions of~$r_1$ and~$r_2$.  This
generalization ability implies that neural models do not need to see
all exponential compositions in order to produce plausible
representations of them. Instead, with appropriate training and model
design, they could uncover and represent the structure underlying
systematically related sentences.


\section{Learning to Disentangle}
\label{sec:learning-disentangle}


While the marking strategy offers substantial benefits in learning
disentangled relation representations, we typically do not have access
to explicit labels indicating which words are helpful for predicting a
specific relation. Nevertheless, the idea of learning representations
specialized for different predictions (albeit with shared parameters)
is general and could potentially alleviate the entanglement problem
for compositional generalization.

Let $[x_1, x_2,...,x_n]$ denote a source sequence. Canonical seq2seq
models like the Transformer \cite{NIPS2017_7181} first encode it into
a sequence of contextualized representations which are then used to
decode target symbols $[y_1, y_2,...,y_m]$ one by one. The same source
encodings are used to predict all target symbols, and are therefore
expected to capture all semantic factors in the input. However,
these could be entangled as demonstrated in our analysis above. To
alleviate this issue, we propose to learn specialized source
representations for different predictions by adaptively re-encoding the source
input at every step of the decoding.

Specifically, at the $t$-th time step, we concatenate the source input
with the previously decoded target and obtain the context for the
current prediction $C_t =
[x_1,x_2,...,x_n,y_1,...,y_{t-1},\mathrm{[PH]}]$ where~$\mathrm{[PH]}$
is a placeholder (e.g.,~a mask token when using a pretrained
encoder). $C_t$~is then fed to a standard encoder (e.g., the Transformer encoder) to obtain the contextualized
representations
$H_t = [h_{t,1},h_{t,2},...,h_{t,n},h_{t,n+1},...,h_{t,n+t}]$:  
\begin{eqnarray}
H_t = f_\texttt{Encoder}(C_t)
\end{eqnarray}
The key difference from the encoder in standard seq2seq models is that
at each time step we adaptively \emph{re-compute} source encodings
\mbox{$H_{t,n}= [h_{t,1},...,h_{t,n}]$} that condition on the newly decoded
target $[y_1,...,y_{t-1}]$. This way, target context informs the
encoder of predictions of interest at each time step. This simple
modification unburdens the model from capturing all source information
through a forward pass of encoding. Instead, based on the simplicity
bias, the model tends to zero in on information relevant for the
current prediction, remaining invariant to irrelevant details, thereby
improving disentanglement. One might argue that the decoder in
standard seq2seq models could also extract specialized information for
each prediction (through the cross attention mechanism). However, it
would fail to do so when working with an entangled encoder that
produces problematic representations for out-of-distribution examples
and breaks down the decoding process.

We propose two strategies for exploiting the target-informed
encoder. Firstly, we use a multilayer perceptron (MLP) to
predict~$y_t$ based on the encoder's output, i.e.,~the last hidden
states~$h_{t,n+t}$:
\begin{eqnarray}
p(y_t|x,y_{<t}) = f_\texttt{MLP}(h_{t,n+t})
\end{eqnarray}
Secondly, we  incorporate the proposed encoder into the standard
encoder-decoder architecture: we take source encodings $H_{t,n}$ and feed
them together with the previous target $[y_1,...,y_{t-1}]$ to a
standard decoder (e.g., Transformer-based) to predict~$y_t$:
\begin{eqnarray}
p(y_t|x,y_{<t}) = f_\texttt{Decoder}(H_{t,n},y_{<t})
\end{eqnarray}
For complex tasks like machine translation, preserving the encoder-decoder architecture is essential to achieving good performance. 

  
We adopt the Transformer architecture to instantiate the encoder and
decoder, however, the proposed method is generally applicable to any
seq2seq model. We maintain separate position encodings for source and
target symbols (e.g., $x_1$ and $y_1$ correspond to the same
position). To differentiate between source and target content, we also
add a source(target) type embedding to all source(target) token
embeddings. Compared to the classical Transformer, our proposal
increases running time from $\mathcal{O}(n^2 + m^2)$ to
$\mathcal{O}(m(n^2 + m^2))$ where~$n$ is input length and $m$~is
output length. Improving the efficiency of our approach is deferred to
future work.

\section{Experiments: Semantic Parsing}
\label{sec:experiments}

In this section, we present our experiments for evaluating the
proposed \textbf{D}isent\textbf{angle}d seq2seq model which we call
\textsc{Dangle}. We refer to the two variants of \textsc{Dangle} as \textsc{Dangle-enc} and \mbox{\textsc{Dangle-encdec}}. We first focus on semantic parsing benchmarks 
which target compositional generalization.  Our second suite of
experiments reports results on compositional generalization for
machine translation.


\subsection{Datasets} 

Our semantic parsing experiments focus on two benchmarks. The first
one is COGS \cite{kim-linzen-2020-cogs} which contains natural
language sentences paired with logical forms based on lambda calculus
(see the examples in Table~\ref{fig:split}).  In addition to the
standard splits of Train/Dev/Test, COGS provides a generalization
(Gen) set that covers five types of compositional generalization:
interpreting novel combinations of primitives and grammatical roles,
verb argument structure alternation, and sensitivity to verb class,
interpreting novel combinations of modified phrases and grammatical
roles, generalizing phrase nesting to unseen depths. 

The former three fall into lexical generalization while the latter two
require structural generalization.  Interpreting novel combinations of
modified phrases and grammatical roles involves generalizing from
examples with PP modifiers within object NPs to PP modifiers within
subject NPs. The generalization of phrase nesting to unseen depths is
concerned with two types of recursive constructions: nested CPs (e.g.,
[\textsl{Mary knows that} [\textsl{John knows} [\textsl{that Emma
  cooks}]$_{\mathrm{CP}}$~]$_{\mathrm{CP}}$~]$_{\mathrm{CP}}$) and
nested PPs (e.g.,~\textsl{Ava saw the ball} [\textsl{in the bottle}
[\textit{on the table}]$_{\mathrm{PP}}$]$_{\mathrm{PP}}$). The
training set only contains nestings of depth 0–2, where depth 0 is a
phrase without nesting. The generalization set contains nestings of
strictly greater depths (3–12). The Train set includes 24,155 examples
and the Gen set includes 21,000 examples.

Our second benchmark is CFQ \cite{keysers2020measuring}, a large-scale
dataset specifically designed to measure compositional
generalization. It contains 239,357 compositional Freebase questions
paired with SPARQL queries. CFQ was automatically generated from a set
of rules in a way that precisely tracks which rules (atoms) and rule
combinations (compounds) were used to generate each example. Using
this information, the authors generate three splits with
\emph{maximum compound divergence} (MCD) while guaranteeing a small
atom divergence between train and test sets. In this dataset atoms
refer to entities and relations and compounds to combinations
thereof. Large compound divergence indicates the test set contains
many examples with unseen syntactic structures. We evaluate our model
on all three splits. Each split consists of 95,743/11,968/11,968
train/dev/test examples.


\subsection{Comparison Models}

On COGS, we trained a baseline \textsc{Transformer}
\cite{NIPS2017_7181} with  sinusoidal (absolute) and relative
position embeddings \cite{shaw-etal-2018-self,
  huang-etal-2020-improve}. We assessed the effect of pretraining on
compositional generalization, by  also fine-tuning \textsc{T5-base}
\cite{Raffel2020ExploringTL} on the same dataset. We created
disentangled versions of these models adopting an encoder-only
architecture (i.e.,~+\textsc{Dangle-enc}). The pretrained version of our
model used \textsc{Roberta} \cite{liu2019roberta}.\footnote{Note that
  we use \textsc{T5-base} instead of \textsc{Roberta} as our
  pretrained baseline on COGS because in initial experiments we found
  that having a pretrained \emph{decoder} is critical for good
  performance, possibly due to the relatively small size of COGS and
  large vocabulary which includes many rare words.}

We also compared with two models specifically designed for
compositional generalization on COGS. The first one is
\textsc{Tree-MAML} \cite{conklin-etal-2021-meta}, a
meta-learning approach whose objective directly optimizes for
out-of-distribution generalization.  Their best performing model uses
tree kernel similarity to construct meta-train and meta-test task
pairs. The second approach is \mbox{\textsc{LexLSTM}}
\cite{akyurek-andreas-2021-lexicon}, an LSTM-based seq2seq model whose decoder is
augmented with a lexical translation mechanism that generalizes
existing copy mechanisms to incorporate learned, decontextualized,
token-level translation rules. The lexical translation module is
intended to disentangle lexical phenomena from syntactic ones.

\begin{table*}[t]
\centering
\begin{tabular}{@{}l@{}ll@{}}
\begin{minipage}[b]{9.1cm}
 \begin{small}
\centering
\begin{tabular}{@{}l|c@{\hspace{.3cm}}c|c@{\hspace{.3cm}}c|c@{\hspace{.3cm}}c|c@{\hspace{.3cm}}c@{}}   \hline
\multicolumn{1}{c}{} & \multicolumn{2}{c}{2} & \multicolumn{2}{c}{3}
& \multicolumn{2}{c}{4} & \multicolumn{2}{c}{5} \\
\multicolumn{1}{@{}c|}{Model} & CP & PP & CP & PP & CP & PP & CP & PP \\
\hline
\textsc{Transformer} (abs) & 3.4 & \hspace*{-1ex}8.9 & 1.2 & 6.6 & 0.8&5.5 & 3.1&8.2  \\
\hspace{.3cm}+\textsc{Dangle-enc} & 11.4 & 5.7 & \hspace*{-1ex}10.3&8.8 & \hspace*{-1ex}14.3&8.6 & \hspace*{-1ex}12.7&\hspace*{-1ex}13.4  \\ 
\textsc{Transformer} (rel) & 0.0 & 0.0 & 0.0 &0.6 & 0.1 & 2.5 & 1.4 & 4.6  \\ 
\hspace{.3cm}+\textsc{Dangle-enc} & \hspace*{-1ex}13.8& \hspace*{-1ex}13.5 & \hspace*{-1ex}18.2&\hspace*{-1ex}19.4 & \hspace*{-1.5ex}24.7&\hspace*{-1ex}31.9 & \hspace*{-1ex}27.2&\hspace*{-1ex}44.3  \\ \hline
\end{tabular}
 \end{small}
\caption{Exact-match accuracy for CP and PP recursion on \textbf{different splits of COGS} (recursion depth with $[2-5]$ range).}
\label{table:results_recursion}
\end{minipage}
& &
\begin{minipage}[b]{6.5cm}
\begin{small}
\begin{center}
\begin{tabular}{@{}l|cccc@{}}
  \hline
 \multicolumn{1}{@{}c|@{}}{Model}  & \multicolumn{1}{@{~}c@{~}}{MCD1} & \multicolumn{1}{@{~}c@{~}}{MCD2} & \multicolumn{1}{@{~}c@{~}}{MCD3}& \multicolumn{1}{@{}c@{}}{Mean} \\
  \hline
  \textsc{T5-11B-mod} & 61.6 & 31.3 & 33.3 & 42.1  \\
  \textsc{HPD} & 72.0 & {66.1 }  & {63.9} & {67.3} \\ 
  \textsc{Roberta} & 60.6 & 33.6 & 36.0 & 43.4  \\ 
   \hspace{.3cm}+\textsc{Dangle-enc} & {78.3} & 59.5 & 60.4 & 66.1  \\  \hline
\end{tabular}
\end{center}
\end{small}
\caption{Exact-match accuracy on \textbf{CFQ}, Maximum Compound
  divergence (MCD) splits.   \label{table:results_cfq}}
\end{minipage}
\end{tabular}
\end{table*}

\begin{table}[t]
\centering
\begin{small}
\begin{tabular}{@{}l|c@{}c@{}c@{}c@{}}
  \hline
 \multicolumn{1}{@{}c|}{Model}  & \multicolumn{1}{c}{OSM} & \multicolumn{1}{c}{CP} & \multicolumn{1}{c}{PP}& \multicolumn{1}{c@{}}{Overall} \\
  \hline
  \textsc{Tree-MAML} & 0.0 & 0.0  & 0.0 & 66.7 \\
   \textsc{LexLSTM} & 0.0 & 0.0  & 1.3 & 82.1 \\  \hline
   \textsc{Transformer} (abs) & 0.0 & 3.4 & \hspace*{-1ex}8.9 & 85.5 \\
  \hspace{.3cm}+\textsc{Dangle-enc}  & 0.0 & 11.4 & 5.7 & 85.9  \\

  \textsc{Transformer} (rel) & 0.0 & 0.0 & 0.0 & 83.3  \\
   \hspace{.3cm}+\textsc{Dangle-enc}  & 0.0 & \hspace*{-1ex}13.8 & \hspace*{-1ex}13.5 & 85.4  \\\hline
    \textsc{T5-base} & 0.0 & \hspace*{-1ex}12.5 & \hspace*{-1ex}18.0 & 85.9 \\   
    \textsc{Roberta} + \textsc{Dangle-enc} & 0.0 & \hspace*{-1ex}\bf{24.6} & \hspace*{-1ex}\bf{34.7} & \bf{87.6}  \\
   \hline
\end{tabular}
\end{small}
\caption{Exact-match accuracy on \textbf{COGS} by 
  type of structural generalization and overall. OSM refers to 
  generalizing from object modifier PPs to subject
  modifier PPs;  CP and PP are recursion depth generalization for
  sentential complements and prepositional phrases.}   

\label{table:results_cogs}
\end{table}

\citet{furrer2020compositional} showed that pretrained
seq2seq models are key to achieving good performance on CFQ.  We
compared against their \mbox{\textsc{T5-11B-mod}} model which obtained
best results among various pretrained models. This is essentially a T5
model with 11B parameters fine-tuned on CFQ with intermediate
representations (i.e.,~SPARQL queries are simplified to be
structurally more aligned to the input for training and then
post-processed to obtain the original valid SPARQL at inference time).
We also built our model on top of \textsc{Roberta} due to the
effectiveness of pre-training on this dataset
(\textsc{Roberta}+\textsc{Dangle-enc}), again adopting an encoder-only
architecture. To tease apart the effect of pretraining and the
proposed approach, we also implemented a baseline that makes use of
the \textsc{Roberta-base} model as the encoder and a vanilla
Transformer decoder.  The Transformer decoder was initialized randomly
and trained from scratch. Finally, we compared against \textsc{HPD}
\cite{NEURIPS2020_4d7e0d72}, a hierarchical poset decoding
architecture which consists of three components: sketch prediction,
primitive prediction, and traversal path prediction. This model is
highly optimized for the CFQ dataset and achieves competitive
performance.

We implemented comparison models and \textsc{Dangle} with fairseq
\cite{ott2019fairseq}; for \textsc{T5-base} we used HuggingFace
Transformers \cite{wolf-etal-2020-transformers}. We provide details on
model configuration, and various experimental settings in the
Appendix.

%
%

\subsection{Results}
Table~\ref{table:results_cogs} shows our results on COGS broken down
by type of structural generalization and overall. All models achieve 0
accuracy on generalizing from PP object modifiers to PP subject
modifiers. We find this is due to a predicate order bias. In all
training examples, ``agent'' or ``theme'' come before preposition
predicates like ``in'', so the models learn this spurious correlation
and cannot generalize to cases where the preposition precedes the
predicate.

Interestingly, a vanilla \textsc{Transformer} outperforms more complex
approaches like \textsc{Tree-MAML} and \textsc{LexLSTM}. We conjecture
the large discrepancy is mostly due to our use of Glove embeddings,
which comparison systems do not use. Pretraining in general
substantially benefits lexical generalization, our
\textsc{Transformer} and \textsc{T5-Base} models achieve nearly
perfect accuracy on all such cases in COGS. An intuitive explanation
is that pretrained embeddings effectively capture common syntactic
roles for tokens of the same type (e.g.,~``cat'' and ``dog'') and
facilitate the generalization of the same decoding strategy to all of
them.  \textsc{Dangle-enc} significantly improves generalization
performance on CP and PP recursion when combined with our base
\textsc{Transformer} and \textsc{Roberta}. 

To further show the potential of our proposal, we evaluated
\textsc{Transformer+Dangle-enc} on additional COGS splits.
Table~\ref{table:results_recursion} shows how model performance
changes with exposure to progressively larger recursion depths. Given
recursion depth~$n$, we created a split by moving all examples with
depth $\leq n$ from Gen to Train set.  As can be seen,
\textsc{Transformer+Dangle-enc}, especially the variant with relative
embeddings, is continuously improving with exposure to additional
training examples. In contrast, vanilla \textsc{Transformer} does not
seem to benefit from additional examples, even when relative position
encodings are used. We can also explain why adding more recursion in
training boosts generalization performance. In the original split,
many nouns never occur in examples with recursion depth 2, which could
tempt the model to exploit this kind of dataset bias for
predictions. In contrast, seeing words in different contexts
(e.g.,~different nesting depth) effectively reduces the possibility of
learning these spurious correlations and therefore improves
compositional generalization.


CFQ results are shown in Table~\ref{table:results_cfq}.
\textsc{Roberta+Dangle-enc} substantially boosts the performance of
\textsc{Roberta-Base}, and is in fact superior to
\textsc{T5-11B-mod}. This result highlights the limitations of
pretraining as a solution to compositional generalization underscoring
the benefits of our approach.  \textsc{Roberta+Dangle-enc} is comparable
to \textsc{HPD} which is a special-purpose architecture highly
optimized for the CFQ dataset. On the contrary, \textsc{Dangle} is
generally applicable to any seq2seq task including machine
translation, as we will show in Section~\ref{mt}.

\subsection{Analysis}
\label{sec:analysis}

\begin{figure*}[t]
\centering
\hbox{\includegraphics[width=0.98\textwidth]{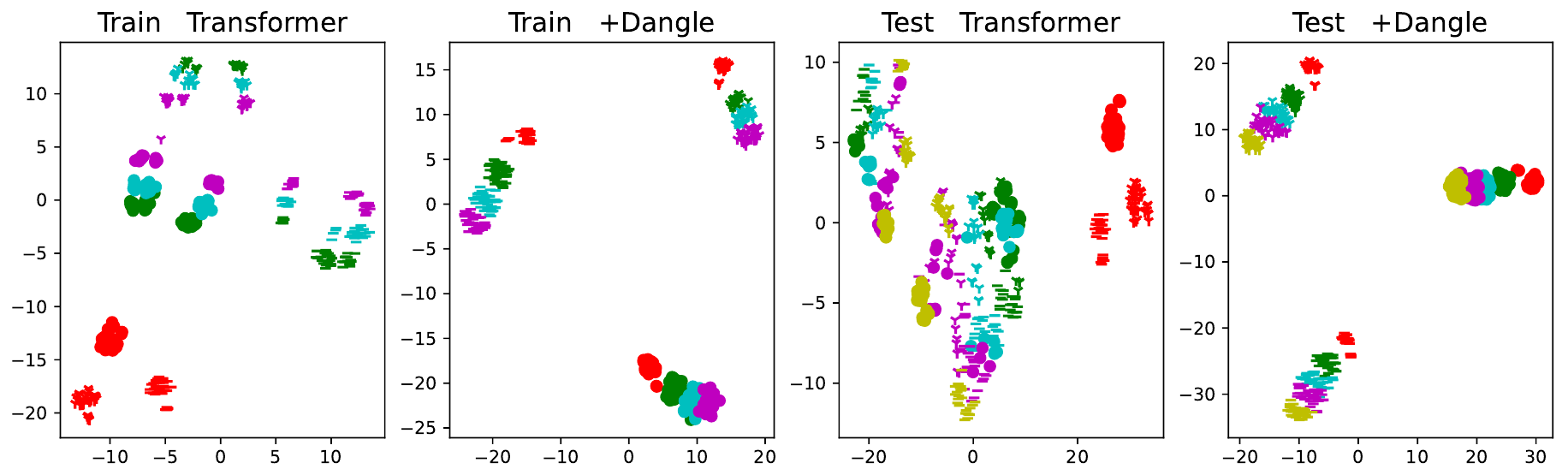}}
\caption{t-SNE visualization of hidden states corresponding to
  predicates ``in'', ``on'', and ``beside'' on training examples with
  PP recursion depth 4 and test examples with PP recursion
  depth~5. Different colors denote different recursion contexts and
  different shape of markers correspond to different predicates.}
\label{fig:t-sne}
\end{figure*}

As discussed in Section~\ref{toy_section}, we hypothesize that a
neural model's inability to perform compositional generalization partly arises from
its internal representations being entangled. To verify this, we
visualize the hidden representations for a \textsc{Transformer} model
with and without \textsc{Dangle}. Specifically, we train both models
on the 4th split of COGS (i.e., data with maximum PP recursion
depth~4) and test on examples with PP recursion depth 5. Then, we
extract the hidden states before the softmax layer used to predict the
preposition predicates ``in'', ``beside'', and ``on'' and use t-SNE
\cite{JMLR:v9:vandermaaten08a} to visualize them.  Ideally, the
representations of these prepositions should be invariant to the
contexts accompanying them so that their prediction is not influenced
by distribution shifts (e.g.,~contextual changes from PP recursion~4
to PP recursion~5).





The visualization is shown in Figure~\ref{fig:t-sne}. Different colors
correspond to different recursion depths while different shape of markers
denote different prepositions (e.g., for a training example like ``NP
in NP in NP in NP in NP'', the hidden states corresponding to the four
``in'' prepositions have the same marker but different colors).  In
training, \textsc{Transformer}'s hidden states within the same
preposition scatter more widely compared to those of \textsc{Dangle},
which implies that its internal representations conflate information
about a preposition's context with itself. In other words,
\textsc{Transformer}'s hidden states capture more context variations
\emph{in addition to} variations corresponding to the predicate of
interest. This in turn causes catastrophic breakdown on the test
examples, where \textsc{Transformer}'s hidden states cannot
discriminate context from predicate information at all. This is in
stark contrast with \textsc{Dangle}, where information about
predicates is preserved even in the presence of unseen contexts.

\begin{table}[t]
\centering
\begin{small}
\begin{tabular}{@{}l@{~}|@{~}c@{~}c@{~}c@{~}|@{~}c@{~}c@{~}c@{}}
  \hline
  & \multicolumn{3}{c|@{~}}{COGS} & \multicolumn{3}{c}{CFQ} \\ 
 \multicolumn{1}{c@{~}|@{~}}{Model}& \multirow{1}{*}{IntraV} & \multirow{1}{*}{InterV} & \multirow{1}{*}{$\downarrow$~R} & \multirow{1}{*}{IntraV} & \multirow{1}{*}{InterV} & \multirow{1}{*}{$\downarrow$~R}\\ 
  \hline
  \textsc{Transformer} & 0.24 & 0.64  & 0.37 & 0.25 & 1.13  & 0.22  \\
  \hspace*{.3cm}+\textsc{Dangle-enc} & 0.19 & 0.73  & 0.26  & 0.01 & 0.52
  & 0.01 \\ \hline
   \textsc{Transformer} & 0.28 & 0.44  & 0.63 & 0.32 & 1.06  & 0.30 \\
  \hspace{.3cm}+\textsc{Dangle-enc} & 0.23 & 0.54  & 0.42 & 0.04 & 0.48  & 0.08 \\ \hline
\end{tabular}
\end{small}
\caption{Entanglement  for \textsc{Transformer} and our approach
  (+\textsc{Dangle-enc}) on COGS and CFQ (for which both models employ a
  \textsc{Roberta} encoder).  Results for training/test set in
  first/second block.  Intra/InterV denotes intra/inter-class variance
  and R~is their ratio.}

\label{table:dis_metric}
\end{table}

We further design a metric to quantify entanglement in neural
representations drawing inspiration from
\citet{pmlr-v80-kim18b}. Their metric assumes 
 the ground-truth factors of a dataset are given, and is applied to
images with one factor fixed and all other factors varying randomly;
if the representation is perfectly disentangled,  the dimension
with the lowest variance should correspond to the fixed factor. Since
in our setting we do not have access to ground-truth factors, we
assume the variable-length target token sequence is the factor of
interest. We also do not need to perform a mapping between neurons and
factors, because their correspondence is hard-coded in seq2seq models
(e.g., a predicate and the hidden units used to predict it).

 For each predicate $y$ occurring in different examples~$e$, we
extract all corresponding representations~$\{\mathbf{v}_{e,y}\}$,
i.e., the last layer of the hidden states used to predict~$y$, and
compute the empirical variance $\Var_e(\mathbf{v}_{e,y}^i)$ for each
$y$; we compute
\mbox{\emph{intra-class}} variance as the average of all predicates'
variances weighted by their respective frequency:
\begin{eqnarray}
V_{intra} &=& \frac{1}{d} \sum_{i=1}^d \E_y \Var_e(\mathbf{v}_{e,y}^i)  \quad  
\end{eqnarray}
where~$d$ is the dimension of hidden states and~$\E$ is the weighted average of
their variances. Intuitively, if the representations are perfectly
disentangled, they should remain invariant to context changes and
intra-class variance should be zero.

We also measure \mbox{\emph{inter-class}} variance by taking the mean
of $\mathbf{v}_{e,y}$ for each predicate~$y$ and then computing the
variance of the means:
\begin{eqnarray}
V_{inter} &=& \frac{1}{d} \sum_{i=1}^d \Var_y \E_e(\mathbf{v}_{e,y}^i)   \quad
\end{eqnarray}
Inter-class variance, on the contrary, should be relatively large for
these hidden states, because they are intended to capture class
variations. The ratio of intra- and inter-class variance collectively
measures entanglement.

As shown in Table~\ref{table:dis_metric},  representations
in \textsc{Dangle} consistently obtain  lower intra- to
inter-class ratios than  baseline models on both COGS and CFQ on
both training and test sets.

\section{Experiments: Machine Translation}
\label{mt}

\begin{table}[t]
\centering
\small
\begin{tabular}{@{}p{7.7cm}@{}} \hline
\multicolumn{1}{c@{}}{\em Training Set} \\ \hline
\cmtt{en:} That winter, Taylor barely moved from the fire. \\
  \cmtt{zh:} 
\begin{CJK*}{UTF8}{gbsn}
  那年冬天， 泰勒几乎没有从大火中挪动过。
  \end{CJK*} \\ \hline
  
\multicolumn{1}{c@{}}{\em Test Set} \\ \hline
\cmtt{en:} That winter, the dog he liked barely moved from the fire. \\
 \mbox{\cmtt{zh:}\begin{CJK*}{UTF8}{gbsn}
 那年冬天，他喜欢的狗狗几乎没有从火堆里挪动过。
 \end{CJK*}} \\ \hline

\end{tabular}
\caption{A training and test example from the CoGnition dataset. The
  test example is constructed by embedding the synthesized novel
  compound ``the dog he liked'' into the template extracted from the
  training example ``That winter, [NP] barely moved from the fire.''.}
\label{fig:mt_split}

\end{table}

\subsection{Dataset} We also applied our approach to CoGnition
\cite{li-etal-2021-compositional}, a recently released realistic compositional
generalization dataset targeting machine translation. This benchmark
includes 216K English-Chinese sentence pairs; source sentences were
taken from the Story Cloze Test and ROCStories Corpora
\cite{mostafazadeh-etal-2016-corpus, mostafazadeh-etal-2017-lsdsem}
and target sentences were constructed by post-editing the output of a
machine translation engine. It also contains a synthetic test set to
quantify and analyze compositional generalization of neural MT
models. This test set includes 10,800 sentence pairs, which were
constructed by embedding synthesized novel compounds into training
sentence templates. Table \ref{fig:mt_split} shows an example. Each
newly constructed compound is combined with 5 different sentence
templates, so that every compound can be evaluated under 5 different
contexts.

\subsection{Comparison Models}
We compared our model to a \textsc{Transformer} translation model
following the same setting and configuration of
\citet{li-etal-2021-compositional}. Again, we experimented with
sinusoidal (absolute) and relative position embeddings.  We adopted
the encoder-decoder architecture variant of our approach (i.e., \textsc{Dangle-encdec}), 
as the encoder-only architecture performed poorly possibly due to the
complexity of the machine translation task.  The number of parameters
was kept approximately identical to the \textsc{Transformer} baseline
for a fair comparison.  All models were implemented using fairseq
\cite{ott2019fairseq}. More modeling details are provided in the
Appendix.

\begin{table}[t]
\centering
\resizebox{\linewidth}{!}{
\begin{tabular}{@{}l|c@{~~}c|c}


  \hline
 Model & $\downarrow~$ErrR$_{\mathrm{Inst}}$ & $\downarrow$~ErrR$_{\mathrm{Aggr}}$ & $\uparrow~$BLEU \\ \hline
  \textsc{Transformer} (abs) & 29.4 & 63.8  & 59.4  \\
  \hspace{.3cm}+\textsc{Dangle-encdec} & 24.4 &  55.5 &  59.7    \\
  \textsc{Transformer} (rel) & 30.5 & 63.8  & 59.4 \\  
  \hspace{.3cm}+\textsc{Dangle-encdec} & \textbf{22.8} & \textbf{50.6}  & \textbf{60.6} \\ 
  
   \hline
\end{tabular}}
\caption{BLEU and compound translation error rates (ErrR) on the
  compositional generalization test set. Subscript $\mathrm{Inst}$ denotes
  instance-wise error rate while $\mathrm{Aggr}$ denotes
  aggregate error over 5 contexts. All results are averaged over 3 random
  seeds. }  

\label{table:results}
\end{table}

\subsection{Results}

As shown in Table~\ref{table:results}, \textsc{+Dangle-encdec} improves over
the base \textsc{Transformer} model by~1.2 BLEU points when relative
position embeddings are taken into account.  In addition to BLUE,
\citet{li-etal-2021-compositional} evaluate compositional
generalization using novel compound translation error rate which is
computed over instances and aggregated over contexts. +\textsc{Dangle-encdec}
variants significantly reduce novel compound translation errors both
across instances and on aggregate by as much as 10 absolute accuracy
points (see first two column in Table~\ref{table:results}). Across
metrics, our results show that \textsc{+Dangle-encdec} variants handle
compositional generalization better than the vanilla.
\textsc{Transformer} model.

\subsection{Analysis}
Two natural questions emerge given the substantial gain achieved
 by \textsc{Dangle} on the compositional generalization (CG) test set:
 (a) Is this gain related to our treatment of the entanglement problem?
 and (b)~How does entanglement manifest itself in machine translation?
 We attempt to answer these questions with an example.

In the CG test set, five new utterances are constructed by embedding
 the novel compound "behind the small doctor on the floor" into five
 sentence templates. In the training set, the phrases ``behind the
 [ADJ] [NOUN]'' and ``the [ADJ] [NOUN] on the floor'' appear
 frequently, but the phrase ``behind the [ADJ] [NOUN] the [ADJ]
 [NOUN]'' is very rare. This poses a serious challenge for the
 baseline encoder-decoder model, which mistakenly translates the
 compound phrase into \mbox{\begin{CJK*}{UTF8}{gbsn} 地板 后面 的 小
 医生\end{CJK*}} (the small doctor behind the floor),
 or \mbox{\begin{CJK*}{UTF8}{gbsn} 地板 上 的 小 医生\end{CJK*}} (the
 small doctor on the floor), or altogether ignores the translation of
 some content words like \mbox{\begin{CJK*}{UTF8}{gbsn} 地板 后面\end{CJK*}}
 (behind the floor). It seems the baseline model cannot simultaneously
 represent the relation between ``behind'' and ``the small
 doctor'' and the relation between ``the small doctor'' and ``the
 floor'', even though the two are conditionally independent. In
 contrast, \textsc{Dangle} generates the correct
 translation \mbox{\begin{CJK*}{UTF8}{gbsn} 地板 上 的 小医生 后
 面\end{CJK*}} in all five contexts. We believe this is due to the
 proposed adaptive encoding mechanism and its ability to decompose the
 representation problem of an unfamiliar compound phrase into
 sub-problems of familiar phrases (i.e, ``behind the small doctor''
 and ``the small doctor on the floor'').

\section{Related Work}
The realization that neural sequence models struggle in settings
requiring compositional generalization has led to numerous research
efforts aiming to understand why this happens and how to prevent
it. One line of research tries to improve compositional generalization
by adopting a more conventional grammar-based approach
\cite{herzig2020spanbased}, incorporating a lexicon or lexicon-style
alignments into sequence models
\cite{akyurek-andreas-2021-lexicon,zheng2020compositional}, and
augmenting the standard training objective with attention supervision
losses \cite{Oren:ea:2020,yin-etal-2021-compositional}. Other work
resorts to data augmentation strategies as a way of injecting a
compositional inductive bias into neural models
\cite{jia-liang-2016-data, Akyurek:ea:2020, andreas-2020-good} and
meta-learning to directly optimize for out-of-distribution
generalization \cite{conklin-etal-2021-meta}. There are
also several approaches which explore the benefits of large-scale
pre-trained language models
\cite{Oren:ea:2020,furrer2020compositional}.

In this work we identify the learning of representations which are not
disentangled as one of the reasons why neural sequence models fail to
generalize compositionally. Disentanglement, i.e., the ability to
uncover explanatory factors from data, is often cited as a key
property of good representations \cite{bengio2013representation}.
For example, a model trained on 3D objects might learn factors such as
object identity, position, scale, lighting, or colour. Several types 
of variational autoencoders \cite{Kingma2014AutoEncodingVB} have been 
proposed for the unsupervised learning of disentangled representations in images
\cite{Higgins2017betaVAELB,pmlr-v80-kim18b,NEURIPS2018_1ee3dfcd}. However, some
of the underlying assumptions of these models have come under scrutiny
recently \cite{pmlr-v97-locatello19a}.

Disentanglement for linguistic representations remains under-explored,
and has mostly focused on separating the style of text from its
content \cite{john-etal-2019-disentangled,
cheng-etal-2020-improving}. In the context of sentence-level
semantics, disentangled representations should be able to discriminate
among lexical meanings and semantic relations between words. We
highlight the entanglement problem in neural sequence models when
trained with explicit factor supervision which, however, does not
cover the entire exponential space of compositions for different
factors. Instead of encouraging disentanglement with some
regularization \cite{Higgins2017betaVAELB,pmlr-v80-kim18b}, we propose
a modification to sequence-to-sequence models which achieves this by
re-encoding the source based on newly decoded target context. It may
be counter-intuitive that we are disentangling by conditioning on more
information, but it is feasible thanks to the inherent
simplicity bias in neural models.

\section{Conclusions}
In this paper we proposed an extension to sequence-to-sequence models
which allows us to learn disentangled representations for
compositional generalization. We have argued that taking into account
the target context makes it easier for the encoder to exploit
specialized information for improving its predictions. Experiments on
semantic parsing and machine translation have shown that our proposal
improves compositional generalization without any model, dataset, or
task specific modification.

\paragraph{Acknowledgments} We thank Chunchuan Lyu,  Bailin Wang, and
the anonymous reviewers for their useful feedback and Yafu Li for his help
with our machine translation experiments. We gratefully acknowledge
the support of the European Research Council (award number 681760).

\bibliography{custom}
\bibliographystyle{acl_natbib}

\appendix

\section{Model Configuration: Semantic Parsing Experiments}


In these sections, we describe the configuration of the models
evaluated in the experiments of Sections~\ref{sec:experiments}
and~\ref{mt}, respectively.

On COGS, the small in-distribution development (Dev) set makes model
selection extremely difficult and non-reproducible. We follow
\citet{conklin-etal-2021-meta} and sample a small subset
from the generalization (Gen) set denoted as `Gen-Dev' for tuning
hyper-parameters. Best hyper-parameters were used to rerun the model
with 5 different random seeds for reporting final results on the Gen
set.  For the baseline \textsc{Transformer}, the layer number of
encoder and decoders are both 2. The embedding dimension is 300. The
feedforward embedding dimension is 512. For
\textsc{Transformer+Dangle}, to maintain approximately identical model
size with the baseline, we used the same embedding dimension and set
the number of the encoding layers to~4. For both models, we
initialized embeddings (on the both source and target side) with Glove
\cite{pennington2014glove}.


On COGS, for the \textsc{Roberta+Dangle} model, we share the target
vocabulary and embedding matrix with the source. On CFQ, we use a
separate target vocabulary; the target embedding matrix is randomly
initialized and learned from scratch.  \textsc{Roberta-base} on CFQ is
combined with a Transformer decoder that has 2 decoder layers with
embedding dimension~256 and feedforward embedding dimension~512. All
hyper-parameters are chosen based on validation performance. On CFQ,
for both \textsc{Roberta-base} and \textsc{Roberta+Dangle}, results
are averaged over 3 randoms seeds.

\section{Model Configuration: Machine Translation Experiments}

We followed the same setting of \citet{li-etal-2021-compositional} and
adopted a \textsc{Transformer} translation model consisting of a
6-layer encoder and a 6-layer decoder with hidden size~512. Each
training batch includes 8,191 tokens at maximum. This model was
trained for 100,000 steps and we chose the best checkpoint on the
validation set for evaluation. Again, we experimented with sinusoidal
(absolute) and relative position embeddings.

We used the same hyperparameters as the baseline model except for the
number of layers which we tuned on the validation set; for relative
position embeddings, the encoder has~4 vanilla source-only Transformer
encoder layers on top of 4 target-informed Transformer encoder layers
(i.e.,~8 encoder layers in all) and the decoder has~4 Transformer
decoder layers; for absolute position embeddings, the encoder has~4
vanilla source-only Transformer encoder layers on top of 2
target-informed Transformer encoder layers and the decoder has~6
Transformer decoder layers. For a fair comparison, we also experimented with 8 encoder layers and 4 decoder layers for the baseline \textsc{Transformer}, and found that it performs similarly to the standard 6-layer architecture.

\end{document}